\pdfoutput=1
\documentclass[11pt]{article}

\usepackage{ACL2023}

\usepackage{csvsimple}
\usepackage{longtable}
\usepackage{enumitem}
\usepackage{booktabs}
\usepackage{numprint}
\usepackage{supertabular}
\usepackage{xurl}
\usepackage{tikz}
\usetikzlibrary{tikzmark}
\usepackage[T1]{fontenc}

\usepackage[utf8]{inputenc}

\usepackage{microtype}

\usepackage{inconsolata}

\usepackage{comment}

\usepackage{times}
\usepackage{latexsym}

\usepackage{microtype}
\usepackage{graphicx}
\usepackage{times}
\usepackage{amsmath}
\usepackage{helvet}
\usepackage{courier}
\usepackage{tabularx}
\usepackage{comment}
\usepackage{pgfplots}
\usepackage{todonotes}
\usepackage{siunitx}
\usepackage{array,multirow}
\usepackage{tabularx,array,booktabs}
\newcolumntype{Y}{>{\RaggedRight\arraybackslash}X} 
\definecolor{red1}{HTML}{F38BA0}
\title{Human-in-the-loop Evaluation for Early Misinformation Detection: \\ A Case Study of COVID-19 Treatments}

\author{Ethan Mendes, Yang Chen, Wei Xu, Alan Ritter\\
  Georgia Institute of Technology \\
  \texttt{\small \{emendes3, yangc\}@gatech.edu \quad \{wei.xu, alan.ritter\}@cc.gatech.edu}
\\}
\begin{document}
\maketitle
\begin{abstract}
We present a human-in-the-loop evaluation framework for fact-checking novel misinformation claims and identifying social media messages that support them.  Our approach extracts check-worthy claims, which are aggregated and ranked for review.  Stance classifiers are then used to identify tweets supporting novel misinformation claims, which are further reviewed to determine whether they violate relevant policies.  To demonstrate the feasibility of our approach, we develop a baseline system based on modern NLP methods for human-in-the-loop fact-checking in the domain of COVID-19 treatments. 
We make our data\footnote{\url{https://github.com/ethanm88/hitl-evaluation-early-misinformation-detection}} and detailed annotation guidelines available to support the evaluation of human-in-the-loop systems that identify novel misinformation directly from raw user-generated content.


\end{abstract}

\section{Introduction}


As many people now get information from social networking websites such as Facebook and Twitter, misinformation has become a serious societal problem. To address this, social media companies have spent billions on content moderation.\footnote{\url{https://www.cnbc.com/2021/02/27/content-moderation-on-social-media.html}}  Prior work on developing natural language processing systems to combat misinformation has mainly focused on various sub-tasks \cite{lee-etal-2021-unifying,guo2022factcheck}, including claim detection \cite{eger-etal-2017-neural,li2022covid}, evidence retrieval \cite{jiang-etal-2020-hover, samarinas-etal-2021-improving, wan-etal-2021-dqn,rami2022natural}, fact verification \cite{aly-etal-2021-fact,wu-etal-2022-cross,jifan2022generating,zihui2022pasta}, stance classification \cite{thorne-etal-2017-fake-stance, conforti-etal-2018-towards, li-etal-2019-rumor-detection}, and fallacy recognition \cite{alhindi2022}. Researchers have also attempted to perform early detection of novel misinformation claims \citep{yue2022contrastive}, as it is crucial for supporting early interventions such as pre-bunking \citep{lewandowsky2021countering}. However, evaluations are often set up automatically using datasets that were retrospectively constructed based on a predefined set of debunked claims.

Recent work by \citet{glockner2022missing} presented convincing evidence that existing NLP fact-checking pipelines are unsuitable for detecting novel real-world misinformation.  They show these systems rely on leaked counter-evidence from news sources that have already fact-checked the claim.
In general, it is unrealistic to assume this type of evidence will be available for new claims that have not yet been widely spread.

\begin{figure*}[ht!]
    \centering
    \includegraphics[width=\textwidth]{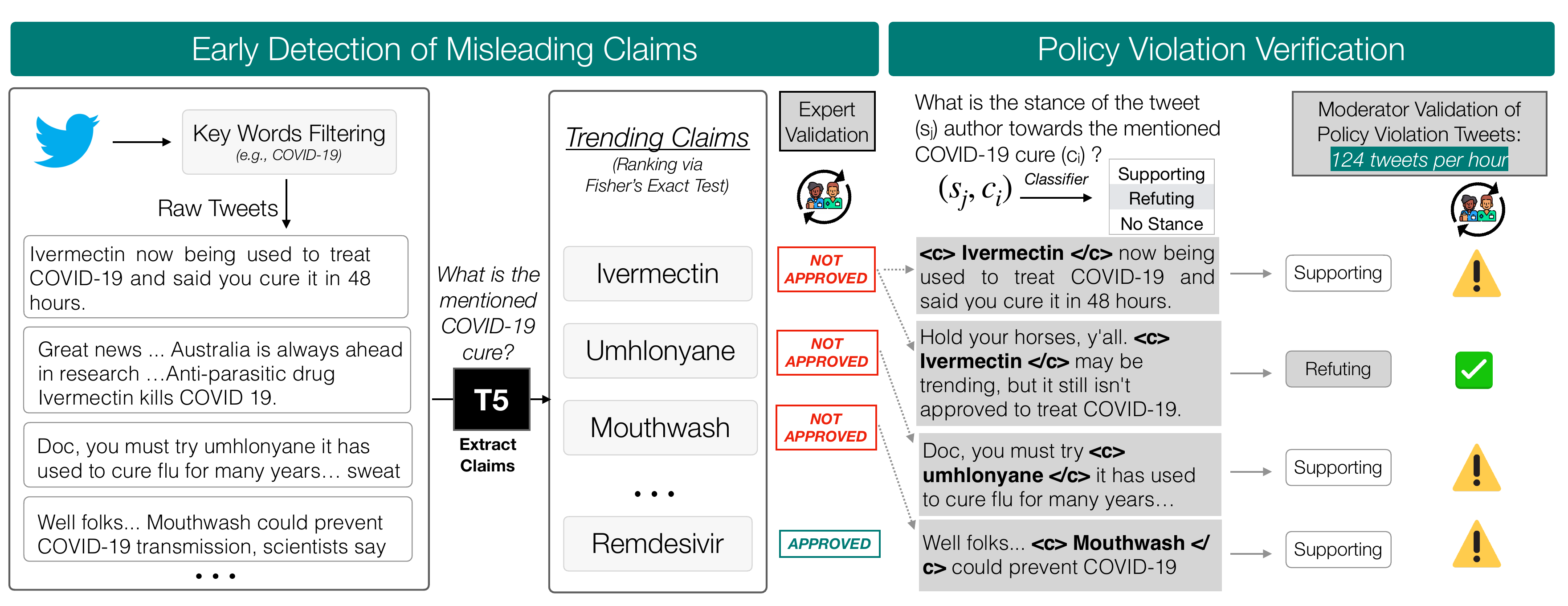}
    \caption{Overview of our human-in-the-loop evaluation framework for early misinformation detection. In stage one (left), a system extracts check-worthy claims directly from raw tweets \textit{in the wild} (rather than retrieving relevant tweets based on provided claims), then aggregates trending claims to be validated by human experts. In stage two (right), the system classifies authors' stances toward false claims and flags tweets for further manual inspection. 
    }
    \label{fig:pipeline}
\end{figure*}

In this paper, we address this challenge by presenting a more realistic human-in-the-loop detection and evaluation framework that can measure a system's capabilities for detecting novel check-worthy claims {\em in the wild} (see Figure \ref{fig:pipeline}). We focus on discovering new, domain-specific claims from raw tweets which are then verified by humans, rather than relying on a pre-defined list of claims that have already been fact-checked for evaluation.  More importantly, we consider not only the \textit{accuracy} but also the \textit{volume}, \textit{relevance}, and \textit{timeliness} of misinformation claims automatically identified by a system, given a collection of raw tweets. We argue this approach provides more realistic experimental conditions because \textbf{(1)} it does not rely on leaked counter-evidence from claims that have already been fact-checked, \textbf{(2)} human expertise is vital in verifying the truthfulness of claims \cite{Nakov2021AutomatedFF,karduni2018can} and \textbf{(3)} it is more effective for humans to check aggregated claims within a specific domain (e.g., claims about the efficacy of COVID-19 treatments), before proceeding to individual social media messages to determine if they violate specific misinformation policies.

We validate our methodology for end-to-end misinformation detection in the domain of COVID-19 treatments.  COVID-19 treatments make an ideal testbed for human-in-the-loop misinformation extraction because Twitter has provided clearly defined policies in this area, which we use as guidelines in a realistic human evaluation of a system's output.\footnote{\tiny \url{https://tinyurl.com/CovidMisinformationPolicy}} 
We evaluate our baseline system with our four defined metrics and find that $18\%$ of the top-$50$ trending claims were actually misleading (\textit{relevance}), $50\%$ of new misleading claims (unapproved COVID-19 treatments) are detected before they are debunked by journalists in a news article (\textit{timeliness}), $65\%$ of tweets flagged constitute policy-violations (\textit{accuracy}), and an average of $124$ policy violations can be confirmed by a human-annotator per hour (\textit{volume}) when using our system.


Our work fills an important gap in the literature, by showing that it is possible to construct a realistic end-to-end evaluation that supports the early detection of novel rumors directly from raw data. Instead of classifying individual tweets as rumorous or not, we extract phrase-level claims that can be aggregated and ranked across a large amount of data and thus can be reviewed more time-efficiently by fact-checkers for human evaluation and for real-world applications. Tweets that are automatically classified as supporting misinformation claims can then be reviewed to determine whether they violate relevant policies.

\section{Related Work}
\label{sec:related_work}

There is a large body of misinformation-related research. Due to space limitations, we only highlight the most relevant work. See also the excellent surveys by \citet{Nakov2021AutomatedFF} and \citet{guo2022factcheck}. 


\subsection{Detecting Check-worthy Claims}
\label{subsec:related_work_checkworthiness}

One of the most related works to ours is the CLEF-2022 CheckThat shared-task \cite{10.1007/978-3-031-13643-6_29}, which evaluates three sub-tasks automatically and separately: \textbf{(1)} determine whether a tweet is worth fact-checking; \textbf{(2)} given a check-worthy claim in the form of a tweet, and a set of previously fact-checked claims, rank the tweets in order of their usefulness to fact-check; and \textbf{(3)} given the text and the title of a news article, determine whether
the main claim it makes is true, partially true, false, or other. In contrast, our experimental setup is more realistic as it operationalizes over a large amount (e.g., millions) of raw tweets and requires span-level extraction to identify the exact claims (e.g., claims about the efficacy of COVID-19 treatments) rather than just ``claims in the form of a tweet'' (e.g., tweets that talk about COVID-19 treatments). We also present an end-to-end human-in-the-loop evaluation of the entire misinformation detection pipeline based on the accuracy, volume, and timeliness of all extracted claims, other than just the automatic intrinsic evaluation of each component separately. 

Similar to CLEF CheckThat, there exist many other prior works that treat claim detection (or rumor detection) as a text classification problem by predicting check-worthiness (or rumourousness) given a tweet or sentence. One representative work is ClaimBuster \cite{10.1145/3097983.3098131} which classifies 20,617 sentences from the U.S. general election debate transcripts as non-factual, unimportant factual, and check-worthy factual. Researchers have also developed other datasets \cite{diggelmann2020climate, 10.1145/3412869,10.1145/3308558.3313618,thorne-etal-2018-fever} and automatic models \cite{10.1145/3308560.3316736, jaradat-etal-2018-claimrank, wright-augenstein-2020-claim}. Another relevant work is by \citet{sundriyal-etal-2022-empowering} which identifies claims as text fragments, such as ``our wine keeps you from getting \#COVID19'' and ``Better alternative to \#DisinfectantInjection''. The evaluations are mostly done automatically over a small fixed set (normally at the scale of 1k$\sim$50k) of annotated tweets or sentences. 


\subsection{Early Rumor Detection} 
As briefly mentioned in \S \ref{subsec:related_work_checkworthiness}, rumor detection is also commonly framed as a text classification task. The standard rumor detection setup \cite{Zubiaga2016LearningRD,derczynski-etal-2017-semeval,10.1145/3070644,gorrell-etal-2019-semeval,kai2020} considers only accuracy without temporal information in the evaluation. More related to our work is a task called early rumor detection \cite{10.1145/2806416.2806651,ma-etal-2017-detect,10.5555/3172077.3172434,10.1145/3132847.3132877,zhou-etal-2019-early,xia-etal-2020-state,Bian2020RumorDO}, which compares classification model's accuracy at different time points and has been extensively surveyed and discussed by \citet{zeng-gao-2022-early}. However, as they pointed out, most existing methods were ``designed with oversimplification'' and evaluated automatically on datasets, such as TWITTER-WEIBO \cite{10.5555/3061053.3061153}, PHEME \cite{Zubiaga2016LearningRD}, and BEARD \cite{zeng-gao-2022-early}, that were constructed retrospectively by collecting social media posts using manually curated search keywords (e.g., names of false treatment) based on a given set of debunked claims (e.g., from \url{snopes.com}). This setup does not measure systems' capability to discover unseen rumors in the wild as our human-in-the-loop evaluation does. In real-world scenarios, what exactly is needed from a misinformation detection system is to automatically figure out what keywords (e.g., names of potential false treatments) to search for -- which we focus on and evaluate in this paper.

%





\subsection{COVID-19 Misinformation Detection}

Given the severity and pervasiveness of the issue, there exists a lot of research (not limited to NLP) about COVID-19 misinformation  \cite{hossain-etal-2020-covidlies,glandt-etal-2021-stance,10.1145/3340531.3412765,SHAHI2021100104,agley2021misinformation,chen-hasan-2021-navigating,biamby-etal-2022-twitter}. The most related work to ours is the CONSTRAINT shared-task \cite{10.1007/978-3-030-73696-5_5} at the AAAI-2021, which considers a binary text classification problem of 10,700 COVID-related tweets about real and fake news, and in particular, the work by \citet{ijcai2022p706} that experimented on this dataset with a human-in-the-loop approach. For each input tweet (e.g., ``Ethylene oxide used in COVID-19 testing swabs changes the structure of the DNA of human body''), \citet{ijcai2022p706} asked crowd workers to write out the main message (e.g., ``Ethylene oxide somehow damages human DNA''), which is then compared with information extracted from COVID-related fack-checking articles and medical papers to help automatic system to predict the tweet's truthfulness. While they prototyped the interesting idea of human-in-the-loop misinformation detection \cite{Shabani2021SAMSHA}, their design is unrealistic to require humans to manually write one sentence per tweet.



\begin{figure*}[hbt!]
    \centering
    \includegraphics[width=\textwidth]{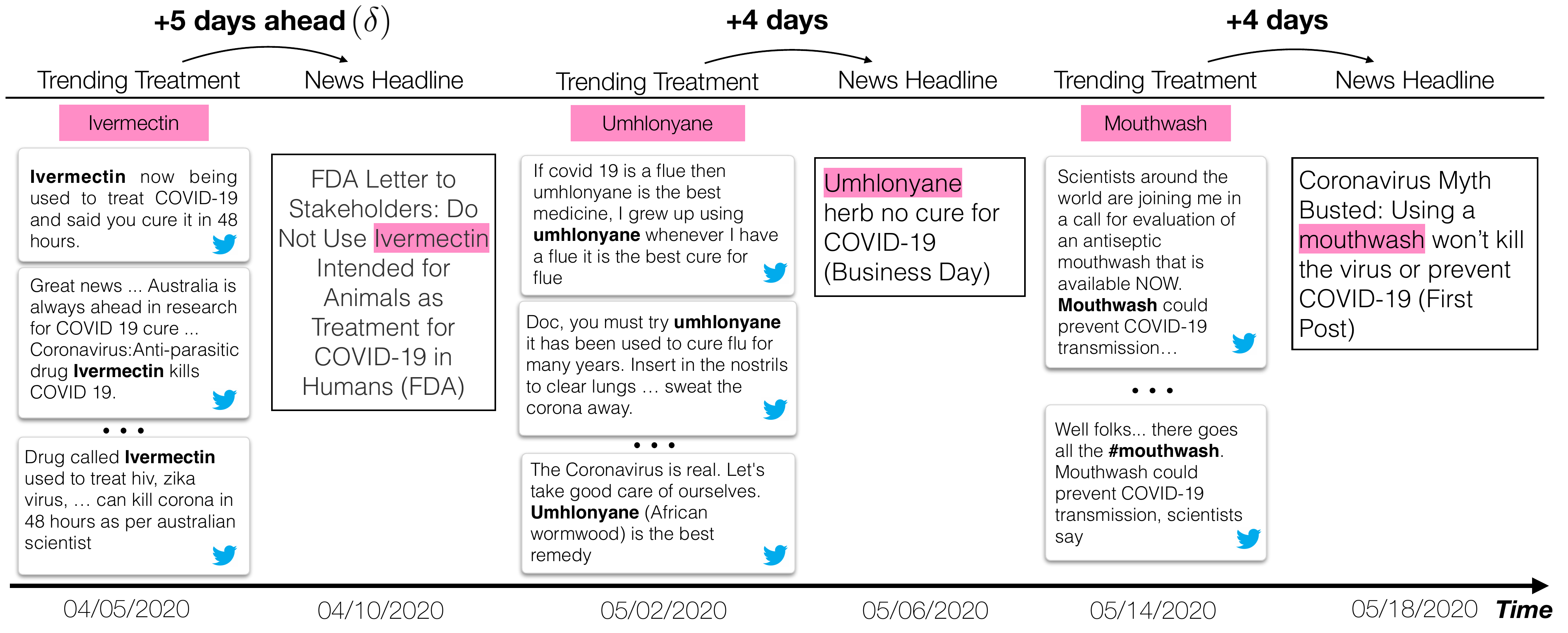}
    \caption{Examples of trending claims regarding notable unapproved treatments detected by our system.  We also show the headline of the news reference identified by our annotators in the mock content moderation experiment.  For each claim, two/three tweets are displayed that were classified as {\sc supporting} from the date that the trend was detected.} 
    \label{fig:demo}
\end{figure*}

\section{Human-in-the-Loop Evaluation Framework}
\label{sec:hitl_eval}
One of the most important functions of a misinformation detection system is to identify new misinformation claims \textit{in the wild}, and in a timely manner. We thus design our evaluation framework to measure not only the \textit{accuracy} but also the \textit{volume}, \textit{relevance}, and \textit{timeliness} of misinformation claims identified by a system, given a large collection of raw tweets (\textit{not} collected based on already debunked claims). See Figure~\ref{fig:pipeline} for an overview of our framework.

\subsection{Early Detection of Misleading Claims}
\label{sec:early_detection}
\paragraph{Problem Definition.}
Given a large set of tweets $\mathbf{T}$, the goal is to automatically discover novel check-worthy claims, each denoted $c_i$, and aggregate a ranked list of claims, $C = [c_1, c_2, ..., c_n]$. 
In this task, we use \textit{trendiness} (e.g. defined with Fisher's Exact Test in \S\ref{subsec:ranking_trending}) as a factor in ranking claims, where a more popular or widely discussed claim will have a greater trendiness.
More formally, a novel check-worthy claim $c_i$ is characterized by $(t_i, z_i, \mathbf{S}_i)$, where $t_i$ is the first time the system identified the claim as trending i.e. when its trendiness first broke a set threshold, $z_i$ is the claim's trendiness score at time $t_i$, and $\mathbf{S}_i \subset \mathbf{T}$ is the set of tweets supporting the claim. A filtering heuristic can also be applied to remove obvious non-misleading claims from consideration (e.g. filtering out claims supporting approved COVID-19 treatments in \S\ref{subsec:ranking_trending}).
\paragraph{Evaluation Metrics.}

A human-in-the-loop evaluation is performed over the top-$K$ trending claims in which annotators verify whether a claim is misleading and, if so, find the earliest news article debunking the claim.
The evaluation is based on two metrics: \textbf{(1)} the percentage of misleading claims in the top-$K$ trending claims (\textit{relevance}) and \textbf{(2)} the number of days (or hours), denoted as $\delta$, between 
$t_i$ and the publication date of the earliest news article (\textit{timeliness}). Figure \ref{fig:demo} visually depicts the application of the $\delta$ metric for COVID-19 treatment misinformation. See $\S$\ref{sec:extrinsic_human_eval} for details about this case study evaluation. 


By annotating at the claim level before the tweet level, we reduce human annotator workload by limiting the total number of tweets evaluated.
This approach also makes our framework more realistic and efficient, allowing for a more thorough and accurate evaluation of misinformation detection systems. In order to facilitate future research in this area, we provide a collection of raw tweets to evaluate and a baseline system to compare against. 
As most existing systems are not available as open-source, the release of our evaluation platform will enable fair and comparable evaluations of these systems. 

\subsection{Policy Violation Verification}
\label{sec:policy_violation}
\paragraph{Problem Definition.}
The objective of this task is to identify tweets within the set associated with a claim $c_i$, $\mathbf{S}_i = \{s_1, s_2, ..., s_{|\mathbf{S}_i|}\}$, that violate a misleading information policy. 
In general, a tweet $s_j$, is likely to violate a policy if it expresses a strong supportive stance towards a claim $c_i$, which was identified as misleading by the human-in-the-loop evaluation process from the prior stage (\S\ref{sec:early_detection}).

\paragraph{Evaluation Metrics.}
To evaluate a system's performance and effectiveness, a human-in-the-loop evaluation is performed on a random sample of $N$ tweets that express a supportive stance toward misleading claims. 
The evaluation is based on two metrics: \textbf{(1)} the \textit{accuracy} of the system in identifying policy-violation tweets and \textbf{(2)} the \textit{volume} in terms of the number of policy violations found per hour by analysts using the system.

To measure the \textit{accuracy} of the system, human annotators assign a score to each tweet in the sample, based on a five-point Likert scale, with 5 corresponding to a clear violation of the policy and 1 representing a clear non-violation. 
We set a threshold of score $\geq 4$ to make a binary policy violation determination.
This scoring scheme allows us to measure the system's accuracy based on the distribution of annotator scores for all tweets in the sample.  

To quantify the \textit{volume} of policy violations identified by analysts using the system, we define the metric \textit{policy-violations per hour} as the number of tweets identified by the annotator containing policy violations, divided by the total number of hours spent by the annotator during the two-stage annotation process ($\S$\ref{sec:early_detection} and \ref{sec:policy_violation}):
\begin{align*}
    \textit{policy violations / hr} = \frac{V}{C \times r_c + T \times r_t} 
\end{align*}

\noindent where $V$ is the number of policy violations found, $C$ and $T$ are the numbers of claims and tweets checked respectively, and $r_c$ and $r_t$ are the average annotation rates for claims and tweets respectively. 

This metric allows us to assess the efficiency of the system in identifying policy-violation tweets and the potential benefits for content moderators using the system.

\section{A Case Study: COVID-19 Treatment Misinformation}
\label{sec:case_study}


To illustrate the usage of our human-in-the-loop evaluation framework outlined in \S\ref{sec:hitl_eval},  we present a case study for COVID-19 misinformation. Specifically, we target Twitter's COVID-19 policy on unapproved treatments, which states that:
\begin{quote}
    ``\textit{False or misleading information suggesting that unapproved treatments can be curative of COVID-19}''
\end{quote}
are grounds for labeling tweets with corrective information~\cite{twitter_2021}.

\subsection{Our System}
\label{sec:system}
In this subsection, we describe the three components of our COVID-19 treatment misinformation detection system: claim extraction, stance classification, and  claim ranking with their task-specific {\em intrinsic} performance.
Later, in $\S$\ref{sec:extrinsic_human_eval}, we present an {\em extrinsic} human-in-the-loop evaluation of the entire system using our defined framework (\S\ref{sec:hitl_eval}). 


\subsubsection{Extracting Check-Worthy Claims}
\label{section:claim_extraction}
\paragraph{Data.}
We train and evaluate our claim extraction models on the human-annotated Twitter COVID-19 event extraction dataset created by \citet{zong2020extracting}, which is collected between 2020/01/15 and 2020/04/26. In this work, we focus on claims of the form ``{\em X is an effective COVID-19 treatment}'', where {\em X} is an extracted span.
We split the provided $1,271$ training tweets in the \textsc{Cure \& Prevention} category into 60\% for training and 15\% for development, and report token-level F1 scores on the 500 tweets used for evaluation in the 2020 W-NUT shared task.\footnote{\url{https://noisy-text.github.io/2020/extract_covid19_event-shared_task.html}} 
\paragraph{Models.}
We develop three approaches, outlined below, to extract claims as a text span from a sentence with a sequence tagging model and a question-answering model~\citep{rajpurkar-etal-2018-squad2.0,du-etal-2021-qa}. Details of training hyperparameters can be found in Appendix~\ref{sec:appendix-claim}. 

\noindent \textbf{(1) Sequence Tagging:} a standard sequence-labeling task with a BIO tagging scheme, where `B' and `I' tags are used to identify treatment tokens. We follow a similar approach as the named entity recognition method used by ~\citet{devlin-2019-bert} and experiment with two pre-trained models, including RoBERTa$_{large}$~\citep{liu2019roberta} and a domain-specific COVID-Twitter-BERT$_{large}$(CT-BERT)~\citep{muller2020CTBERT}. 

\noindent\textbf{(2) Question-answering (QA):} we treat the claim extraction as a question-answering task and apply the SQuADv2.0~\citep{rajpurkar-etal-2018-squad2.0} formulation as some tweets may not include relevant claims (similar to unanswerable questions). 
  We experiment with two approaches:
    a span-prediction model that predicts start and end positions for answer spans in context using RoBERTa/CT-BERT as the encoder, in addition to a text-to-text model that generates answers using T5$_{large}$~\citep{raffel2020exploring,wang-lillis-2020-ucd}. The question template for extracting treatments discussed in tweets is ``What is the mentioned COVID-19 cure?''. 
    
    \noindent\textbf{(3) QA-Pretraining:} it has been shown that intermediate task pre-training can yield further gains for low-resource target tasks~\citep{pruksachatkun-etal-2020-intermediate,poth-etal-2021-pre}. We thus experiment with pre-training QA models on the SQuADv2.0 dataset before fine-tuning on the claim extraction dataset.
\paragraph{Intrinsic Evaluation.} 
Table~\ref{tab:claim_extraction} shows the claim extraction results on the COVID-19 treatment dataset. 
We observe QA models outperform tagging models across encoders.
RoBERTa outperforms the domain-specific encoder (CT-BERT) for QA extraction.
However, after QA pre-training, CT-BERT improves from 53.1 to 63.8 F$_1$ and outperforms RoBERTa by 2.3 points.
Finally, as the T5 model achieves the best F$_1$ regardless of QA pre-training, we use T5$_{\text{SQuADv2 Pre-train}}$ as our final claim extraction model. 

\begin{table}[t!]
    \centering
    \footnotesize
    \begin{tabular}{llr}
         \toprule
        \textbf{Approach} & \textbf{Model} & \textbf{$F_1$} \\
        \midrule
        \multirow{2}{0pt}{Tagging} & RoBERTa & 50.3\\
        & CT-BERT & 51.2\\
        \midrule
        \multirow{3}{0pt}{QA} 
        &RoBERTa & 61.5\\
        &CT-BERT & 53.1\\
        &T5 & 63.7\\
        \midrule
        \multirow{3}{0pt}{QA$_{\text{Pre-train}}$}  &RoBERTa$_{\text{SQuADv2 Pre-train}}$ & 59.9\\
        &CT-BERT$_{\text{SQuADv2 Pre-train}}$ & 63.8\\
        & T5$_{\text{SQuADv2 Pre-train}}$ & \textbf{63.9}\\
        
         \bottomrule
    \end{tabular}
    \vspace{-5pt}
    \caption{Token-level F$_1$ scores for claim extraction experiments on COVID-19 treatment dataset.}
    \label{tab:claim_extraction}
\end{table}

\subsubsection{Task-Specific Stance Classification}
\label{sec:stance}
\paragraph{Data.}
Due to the lack of datasets for COVID-19 treatment stance, we annotate a new dataset for our evaluation. To collect relevant tweets we tracked the keywords ``cure'', ``prevention'', ``virus'', and ``COVID-19'' 
from November 2020 to December 2020 
using the Twitter API.
We collected $1,055,559$ tweets and claims extracted using our model ($\S$\ref{section:claim_extraction}).
Out of $97,016$ tweets for which a treatment was able to be extracted, we randomly sample $2,000$ tweets to annotate the author's stance on the effectiveness of the treatment.
We paid crowd workers on Amazon MTurk to annotate our data.
Each task consists of a tweet with a highlighted treatment.
We asked workers to determine the author's stance towards the treatment and select among three options ({\sc Supporting}, {\sc Refuting}, or {\sc No Stance}). We decided not to include additional options for irrelevant and sarcastic tweets due to poor annotator agreement in pilot experiments.
Each tweet is labeled by $5$ independent crowd workers. Workers were paid $\$0.20/$HIT, which roughly equates to $\$8.50/$hour.
A screenshot of the annotation interface is provided in Figure \ref{fig:annotation_task} and dataset statistics are summarized in Table~\ref{tab:dataset_distribution}.

\paragraph{Quality Control.}
During the annotation process, we monitored the quality of workers' annotations using their agreement with each other and split the data into 10 batches (200 tweets each) to detect poor annotations in the early stages.
We calculate the annotation agreement of each worker against the majority of 5 workers.
If the worker's agreement is less than 0.75 for a {\sc Supporting} annotation based on a majority vote, we do not allow them to participate in the subsequent annotation batch.
Across all annotations, we find a 0.65 value of Cohen's $\kappa$ \cite{artstein-poesio-cohen-kappa} for the inter-annotator agreement between workers.
The distribution of the dataset based on the majority vote of workers is shown in Table \ref{tab:dataset_distribution}. 
In the case that there is no majority annotation for a given tweet i.e. the $5$ individual annotators are split $2/2/1$ among the three annotation types, we assign the tweet a default annotation of {\sc No Stance}.
We randomly split our annotated dataset of 2,000 tweets into a training set of 1,200 tweets, a development set of 400, and a test set of 400. 
\begin{table}[t!]
    \centering
    \small
    \begin{tabular}{lr}
         \toprule
         \textbf{Majority Annotation} &  \textbf{\#Tweets}\\
        \midrule
         {\sc Supporting} & 743\\ 
         
         {\sc Refuting} & 631\\ 
         
         {\sc No Stance} & 400 \\ 
         
         {No Consensus ({\sc No Stance})} & 226 \\ 
         \midrule
         \textbf{Total} & 2000 \\ 
        \bottomrule
    \end{tabular}
    \vspace{-5pt}
    \caption{Distribution of annotated COVID-19 treatment stance dataset.}
    \label{tab:dataset_distribution}
\end{table}

\paragraph{Models.}
Using the annotated corpus, we develop classifiers to detect the author's stance toward a treatment. 
Specifically, given a claim and a tweet $(c_i, s_j)$, our goal is to predict the author's stance $m_i\in \{${\sc Supporting}, {\sc Refuting}, or {\sc No Stance}$\}$. We experiment with three models including a baseline neural bag-of-words (NBOW) model, RoBERTa$_{large}$~\citep{liu2019roberta} and a COVID-Twitter-BERT$_{large}$ (CT-BERT)~\citep{muller2020CTBERT}.
To indicate the position of the claim in the input, we use relative position encoding (RPE)~\citep{shaw-etal-2018-self} for the NBOW model. For the pre-trained language models, we add special markers around the claim following the best-performing model from \citet{soares2019re}, [{\sc Entity Markers - Entity Start}].
Details of training hyperparameters are in Appendix~\ref{sec:appendix-stance}.

\paragraph{Intrinsic Evaluation.}
Table \ref{tab:evaluation} presents the results for NBOW, RoBERTa, and CT-BERT.
We observe that CT-BERT outperforms all other models, with an $F_1$ score of 66.7.
Generally, we find that these models performed best in classifying tweets with {\sc Supporting} stance and worst on tweets with a {\sc No Stance} label. This latter result is possibly due to annotators classifying those tweets that might be irrelevant to the task as {\sc No Stance}.
\begin{table}[t!]
    \centering
    \footnotesize
    \begin{tabular}{lccc}
         \toprule
         \textbf{Stance} & NBOW & RoBERTa & CT-BERT\\
        \midrule
         {\sc Supporting} & 59.8 & 70.0 & \textbf{74.9}\\
         {\sc Refuting} & 32.8  & 61.9 & \textbf{70.6} \\ 
         {\sc No Stance} & \textbf{56.0} & 45.9 & 54.7\\ 
         \midrule
         Aggregate $F_1$ & 49.5 & 59.3 & \textbf{66.7} \\
         \bottomrule
    \end{tabular}
    \vspace{-5pt}
    \caption{F$_1$ scores on stance classification dataset for classifying author's stance towards the extracted claim.}
    \label{tab:evaluation}
\end{table}
\subsubsection{Ranking of Trending Claims}
\label{subsec:ranking_trending}
Following Twitter's COVID-19 misinformation policy violation guidelines, we focus on tweets that advocate the efficacy of an unapproved treatment.
Thus, we filter out tweets that mention common approved treatments listed in Table~\ref{tab:approved_treatements} in Appendix~\ref{appendix-approved_treatments}, which are prepared according to authorities and news agencies including the CDC and NYT.
Upon determining stance on the filtered set, we only consider tweets with a {\sc Supporting} stance towards the effectiveness of the extracted treatment.
Finally, we remove near-duplicates and cluster the remaining extracted treatments based on word overlap (Jaccard similarity) to enable treatment-level decision-making similar to \citet{basu-2013-powergrading}.
\paragraph{Ranking Claims.} For the claims (treatments) in each cluster, we count the number of tweets mentioning the claim both daily and cumulatively.
Based on these counts, we compute the claim's $p$-value on a given date using the one-tailed Fisher's Exact Test \citep{Fisher2010OnTI} which has been shown to be effective in rare event detection~\citep{moore-2004-rare-events,johnson2007improving,ritter2011data}. A claim's $p$-value is a measure of its trendiness denoted $z_i$ (\S\ref{sec:early_detection}) by which it is ranked relative to other claims.

\paragraph{Detecting Novel Claims.} Based on the results of Fisher's Exact Test, our system automatically detects novel trending claims and flags them for manual inspection by content moderators.
A claim is considered as newly trending if it's $p$-value is less than a preset significance threshold ($\alpha$-level) and it has never broken this threshold previously (further details in Appendix~\ref{sec:appendix-ranking}). Using the notation from \S\ref{sec:early_detection}, if the claim ($c_i$) found to be newly trending on date $t_i$ is judged to be misleading by a human moderator, our system then provides a list of individual tweets ($\mathbf{S}_i$) that {\sc Support} the misleading claim for manual inspection. 


\subsection{Human-in-the-Loop Evaluation for Detecting COVID-19 Misinformation}
\label{sec:extrinsic_human_eval}
In this section, we evaluate the system outlined in \S \ref{sec:system} using the human-in-the-loop evaluation methodology we define in \S \ref{sec:hitl_eval}. We follow the same procedure described in $\S$\ref{sec:stance} to prepare a new dataset containing $14,741,171$ tweets for large-scale evaluation.
We then extract treatments using our QA-based claim extractor and apply our stance detection model to classify authors' stance towards each treatment. After removing tweets without an extracted treatment, the resultant evaluation corpus consists of $1,905,424$ tweets.

\subsubsection{Early Detection of Misleading COVID-19 Treatments}
\label{subsec:early_detection}
We first evaluate the ability of our system to detect newly trending misleading COVID-19 treatment claims and report metrics measuring \textit{relevance} and \textit{timeliness} as defined in \S\ref{sec:early_detection}.
\paragraph{Data Preparation and Human Evaluation.}
We set aside tweets collected from 2020/03/01 to 2020/03/31 (1-month time-frame) to serve as an initial base of historical data to compute cumulative counts for detecting novel trending claims using the Fisher's Exact Test. Newly trending treatments are then identified during the time period of 2020/04/01 to 2022/05/05 (2-year time-frame) using the methodology described in $\S$\ref{subsec:ranking_trending}. The top $300$ treatments are selected based on $p$-values from Fisher's Exact Test which equates to a significance level of $\alpha=1.15\mathrm{e}{-6}$.

We employ two in-house annotators, who act as mock content-moderators, to evaluate these 300 treatments and determine \textbf{(1)} if the extraction is a treatment \textbf{(2)} if the treatment is unapproved and \textbf{(3)} the earliest publication date of a news article debunking the treatment as effective for COVID-19 using the Google News engine. Appendix~\ref{appendix-guidelines} contains further details about the annotation process.

Out of the $300$ treatments, $100$ were annotated by both annotators to determine inter-rater agreement on task \textbf{(2)}, which was $0.87$ as given by Cohen's $\kappa$. On average, it took $89.7$ seconds to complete each treatment annotation.

\begin{figure}[htb!]
\includegraphics[width=0.48\textwidth]{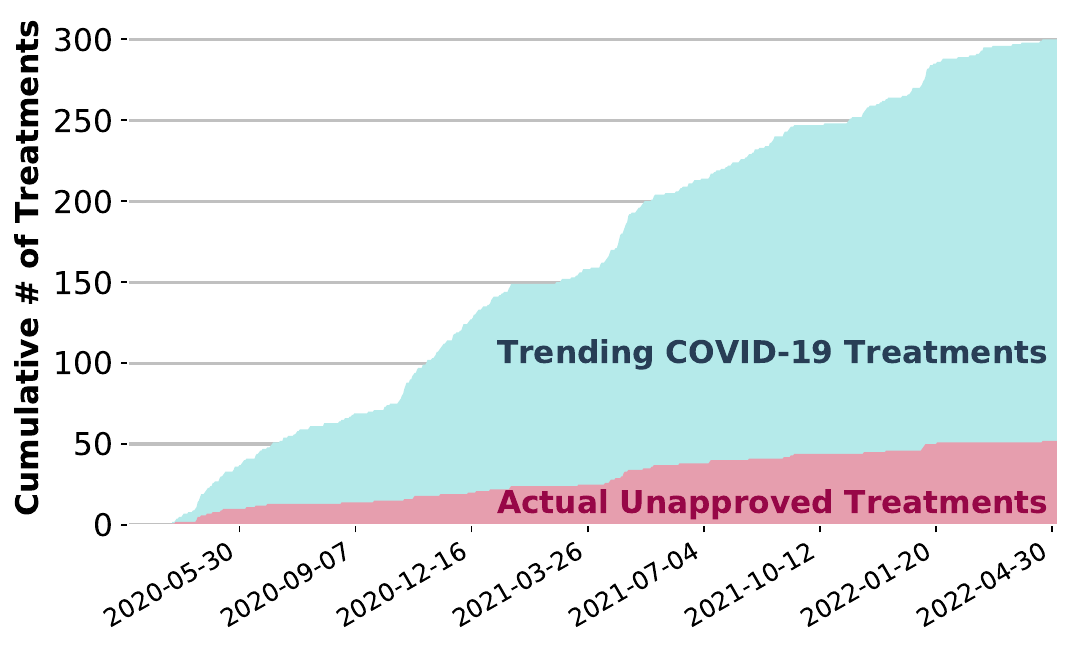}
\caption{Cumulative number of potential and actual unapproved treatments detected.}
\label{fig:cumulative_trends}
\end{figure}

\paragraph{Results on Early Detection of Misleading Claims.}
In terms of \textit{relevance}, Table~\ref{tab:unapproved_trending} shows the percentage of the top 5/50/100 trending treatments ranked by $p$-value that were determined to be unapproved, and Figure~\ref{fig:cumulative_trends} shows the cumulative number of potential unapproved trends identified over time along with the total number of trending treatments. To evaluate the \textit{timeliness},
we calculate $\delta$, which measures the number of days our system detects misleading claims before a debunking news article is published (\S\ref{sec:early_detection}). We find that our system is able to detect $50\%$ of rumors before the publication date of the relevant news article ($\delta \ge 0$), with the median $\delta$  being $21$ days. Figure~\ref{fig:demo} shows three notable unapproved treatment examples from early in the pandemic with their relevant news article and $\delta$ values.
\begin{table}[htbt!]
    \centering
    \small
    \vspace{-5pt}
    \begin{tabular}{lc c c }
         \toprule
          & \textbf{Top 5} & \textbf{Top 50} & \textbf{Top 100} \\
        \midrule
         \% unapproved & 60.0 & 18.0 & 14.0  \\ 
         \bottomrule
    \end{tabular}
    \vspace{-5pt}
    \caption{Percentage of top 5/50/100 trending treatments based on $p$-value that were classified as unapproved in the annotation process.}
    \label{tab:unapproved_trending}
\end{table}

\subsubsection{Identifying COVID-19 Policy Violations.}
\label{subsec:identifying_pv}
In addition to detecting novel rumors online, we also evaluate the ability of our system to identify tweets that violate Twitter's misleading information policy and report metrics measuring \textit{accuracy} and \textit{volume} as defined in \S\ref{sec:policy_violation}.

\paragraph{Data Preparation and Human Evaluation.}
For each of the $40$ treatments identified as unapproved in the previous experiment, we randomly sample $10$ tweets that have {\sc Supporting} stance towards the claim.
Near duplicate tweets were identified and removed leaving $361$ unique tweets.
The two in-house annotators then assign a score to each tweet based on a five-point Likert scale, with 5 corresponding to a clear violation of the policy and 1 representing a clear non-violation (See score descriptions in Table~\ref{tab:likert_descriptions} in Appendix~\ref{appendix-guidelines}).

To investigate the quality of annotations, we compute agreement on $206$ tweets using ordinal Krippendorff's $\alpha$~($0\leq\alpha\leq1$)~\citep{krippendorffalpha2011}.\footnote{We use  Krippendorff’s $\alpha$ instead of Cohen's $\kappa$ to measure inter-rater agreement here because it is applicable to ordinal annotation data.} We find that annotators agreed moderately with Krippendorff's $\alpha=0.54$, indicating fair agreement.
On average, it took the annotators $16.1$ seconds to annotate each tweet.

\begin{figure}[bt]
\centering
    \includegraphics[width=0.4\textwidth]{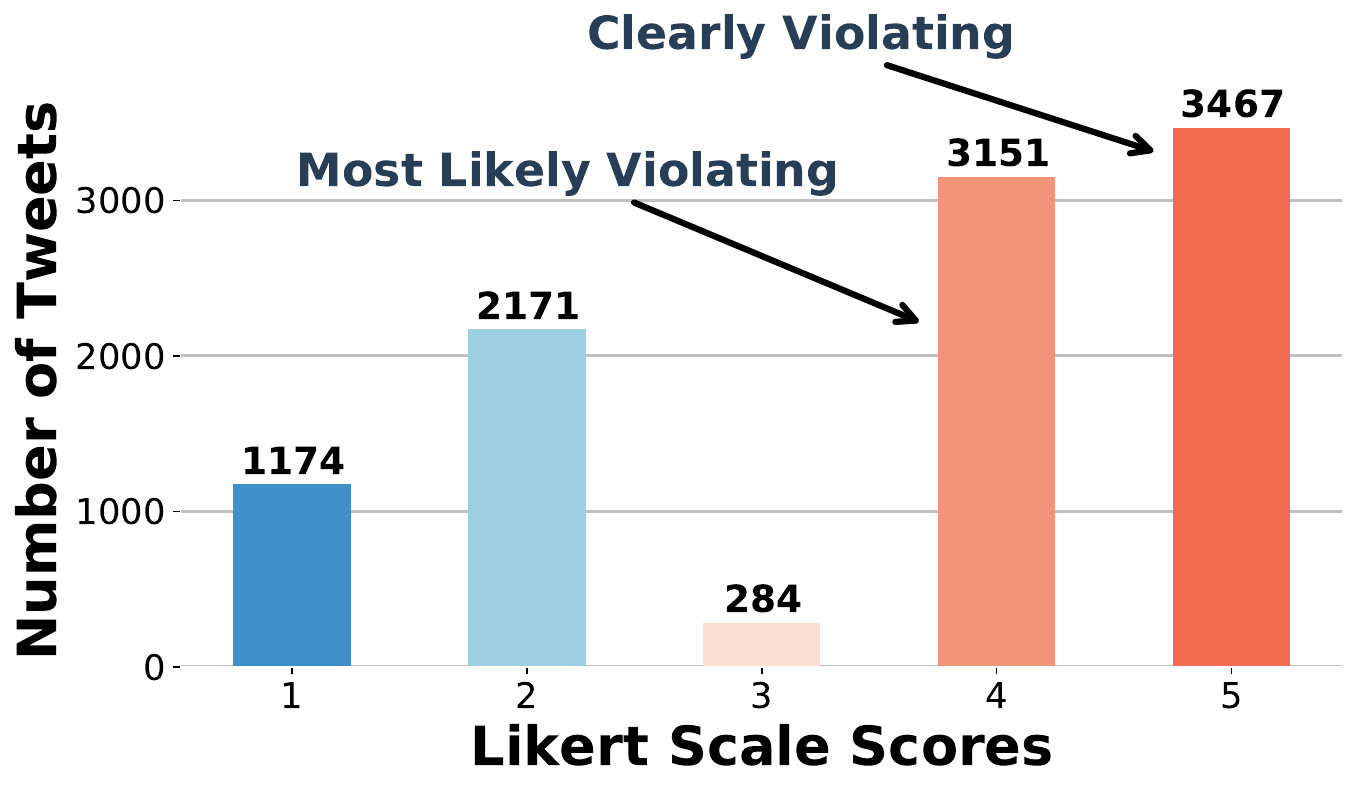}
    \caption{Expected distribution of Likert score annotations on full set of tweets mentioning one of the $40$ treatments investigated.}
    \label{fig:policy_violations}
\end{figure}

\paragraph{Results on Identifying Policy Violations.}
In Figure~\ref{fig:policy_violations}, we find that $65\%$ (\textit{accuracy}) of tweets had scores indicating that it was either likely or clearly violating the policy with an average score of 3.54 out of 5.
Figure~\ref{fig:policy_violations} presents the estimated distribution of Likert scores over 10,246 tweets using the 10 annotated tweets sampled for each treatment and extrapolated to all tweets mentioning the treatment.
In terms of \textit{volume}, we estimate that an annotator can identify approximately $124.2$ policy violations per hour with our system, based on the average annotation rate of the $300$ treatments and the same extended set of tweets, where a Likert score of 4 or 5 constitutes a policy violation. See Appendix \ref{sec:appendixcalculation} for a full calculation of this statistic.

\section{Conclusion}
In this work, we present a novel end-to-end human-in-the-loop evaluation framework for the early identification of novel misinformation on social media from raw tweets. 
Unlike previous evaluation frameworks, our methodology captures the interplay between the system and human content moderators while also providing realistic metrics for early misinformation detection.
We validate our misinformation detection framework for claims in the domain of COVID-19 treatments.
By aggregating and ranking structured representations of claims, and relying on human fact-checkers to review trending claims, our system is able to detect 50\% of misleading claims earlier than the news. 
\pagebreak
\section{Limitations}

While our approach does require domain-specific information extraction models to extract structured representations of novel misinformation claims for easy aggregation and review, there is significant prior work on event extraction that can be adapted to extract check-worthy claims \citep{ritter2012open,luan2019general,du-cardie-2020-event}. Furthermore, we argue content moderators or fact-checkers are likely to be more effective when focusing on one claim type at a time (e.g. COVID-19 treatments, election integrity, vaccine effectiveness, etc.), rather than reviewing a mixture of claims on multiple topics.


Our COVID-19 case study also makes use of ``mock'' content moderators, rather than employees or contractors working for social media companies or fact-checking websites. However, we believe this methodology still provides valuable insight that would not be publicly available otherwise, as social media companies do not currently publish extensive details about their content moderation processes\footnote{\url{https://www.nytimes.com/2022/05/19/business/twitter-content-moderation.html}} and fact-checking websites vary widely in policy and have been shown to provide inconsistent claim classification \cite{marietta-etal-2015-fact-check}. Some prior user studies \cite{10.1145/3242587.3242666,doi:10.1073/pnas.1806781116,Shabani2021SAMSHA} have also shown laypeople (e.g., Amazon Mechanical Turk workers) can be good at judging the veracity of claims or reliability of news articles. 

As of late November 2022, Twitter has suspended enforcement of its COVID-19 misleading information policies such as the one we target in this paper.\footnote{\url{https://apnews.com/article/twitter-ends-covid-misinformation-policy-cc232c9ce0f193c505bbc63bf57ecad6}} However, per the Associated Press article, one of the possible reasons for the suspension was that Twitter has ``struggled to respond to a torrent of misinformation about the virus'' with many ``bogus claims about home remedies'' still on the site despite the previous enforcement of policies. While we do not have details about the internal automated systems Twitter has in place to assist with content moderation, an end-to-end early detection system might have helped stem the spread of misinformation on the platform. Additionally, despite the lack of official policy enforcement, our system can still be used by third-party fact-checking websites or researchers to measure and report misinformation on Twitter. Finally, the main goal of our work is not to create a system for COVID-19 misinformation detection but rather to propose a framework that allows for a fair and realistic evaluation of  early misinformation detection systems in any domain.

\section{Broader Impact and Ethical Considerations}
We release our corpus of tweets annotated with stance, and our dataset of trending misinformation claims under Twitter's Developer Agreement,\footnote{\url{https://developer.twitter.com/en/developer-terms/agreement-and-policy}} which grants permissions for academic researchers to share Tweet IDs and User IDs (less than 1,500,000 Tweet IDs within 30 days) for non-commercial purposes, as of October 10th. 2022.

Our system is designed for research purposes and may contain unknown biases towards demographic groups or individuals \citep{sap2019risk}.  Further investigation into systematic biases should be conducted before our models are deployed in a production environment.

We believe this study helps shed light on how NLP tools developed to help combat online misinformation might be used in a real content moderation workflow. We hope this will encourage future research on human-in-the-loop systems and help shape the design of new tasks and datasets in this area.  We believe it is beneficial for some research on combating misinformation to take place outside of social media companies to provide an unbiased view of the challenges involved in fighting online misinformation.

\section*{Acknowledgments}
We thank anonymous reviewers for their helpful feedback on this work. We also thank Chase Perry and Andrew Duffy for their help with annotations and human evaluation. This research is supported in part by the NSF (IIS-2052498), ODNI and IARPA via the BETTER and HIATUS programs (2019-19051600004, 2022-22072200004). The views and conclusions contained herein are those of the authors and should not be interpreted as necessarily representing the official policies, either expressed or implied, of NSF, ODNI, IARPA, or the U.S. Government. The U.S. Government is authorized to reproduce and distribute reprints for governmental purposes notwithstanding any copyright annotation therein.

\bibliography{custom}
\bibliographystyle{acl_natbib}

\newpage
\appendix
\section{Annotation Guidelines for the Human-in-the-Loop Evaluation}
\label{appendix-guidelines}
\paragraph{Early Detection of Misleading COVID-19 Treatments.}
Given a list of trending claims (e.g., COVID-19 treatments), annotators are required to determine \textbf{(1)} if the extraction is a treatment \textbf{(2)} if the treatment is unapproved and \textbf{(3)} the earliest publication date of a news article they can find that debunks the treatment as effective.
Annotators query the Google News engine with a query in the form of \textit{``[treatment] cures COVID-19''} and sort by date to find the earliest published news article starting from 2020/04/01 that debunks the treatment as effective against COVID-19. Treatments are only considered to be unapproved if the annotators can identify a reputable news source as a reference. Table~\ref{tab:guidelines_early_detection} shows the annotation questions and guidelines as they appeared to annotators during the human-in-the-loop evaluation for this task. Note that March 1st, 2020 was used as the starting date for the article search because it was the earliest date for which we had tweet data.

\begin{table*}[t!]
    \centering
    \small
    \begin{tabularx}{0.9\textwidth}{p{0.25\linewidth} | p{0.60\linewidth}}
        \toprule
         \textbf{Question} & \textbf{Annotation Guidance}\\
        \midrule
        Should the extraction be considered for consideration of a new treatment?   & 
        \setlist{nolistsep}
        \begin{itemize}
            \item Answer ``Repeat'' if the same trend has been seen previously.
            \item Answer``Approved'' if treatment should have been marked as approved based on the approved trending list or is otherwise an obvious valid treatment.
            \item Answer ``Unsure'' if the treatment was in clinical trials at the time or you are otherwise unsure if the treatment is a valid cure.
            \item Answer ``Not a Treatment'' for extraction errors, preventative measures, etc.
            \item Answer ``General health advice" for any general strategies for staying healthy such as ``hand washing'', ``exercise'', ``hygiene'', etc.
            \item Answer ``Unapproved'' otherwise.
        \end{itemize}\\
        \midrule
        What is the date of publication of the earliest article stating that it is misinformation that this treatment cures COVID-19?   & 
        \setlist{nolistsep}
        \begin{itemize}
            \item Answer ``NA'' if you answered ``Approved'', ``Unsure'', ``Repeat'' in the previous question or if no such article can be found in the time frame of [03/01/2020 - Present]. Otherwise, provide the date of the earliest article found.
            \item \textbf{How to search for articles:}
            \begin{enumerate}
                \item If X is the treatment, search: ``X cures COVID-19'' on the News tab in Google - correct any obvious misspellings - try some obvious variations if do not find appropriate results (e.g. ``hcq'' and ``hydroxychloroquine'')
                \item Set date range to [03/01/2020 - Present] (or narrow the date range i.e. month by month range if too many search results)
                \item Select the option to order results by date
                \item Go to the last page in search results (earliest)
                \item Find the earliest article that debunks the claim - answer ``NA'' if no appropriate article is found
            \end{enumerate}
            \item Make sure to verify the correct date of the article publication from the article webpage as the date on google news is not always updated and reliable
        \end{itemize} \\
        \midrule
        What is a link to the article?   & 
        \setlist{nolistsep}
        \begin{itemize}
            \item Article URL or "NA" is the answer to the previous question was "NA"
        \end{itemize} \\
        
    \bottomrule
    \end{tabularx}
    \caption{Human evaluation question and guidelines provided to annotators during the early detection of misleading COVID-19 claims task}
    \label{tab:guidelines_early_detection}
\end{table*}

\begin{table*}[t!]
    \centering
    \small
    \begin{tabularx}{0.9\textwidth}{p{0.25\linewidth} | p{0.6\linewidth}}
        \toprule
         \textbf{Question} & \textbf{Annotation Guidance}\\
        \midrule
        Is the tweet a duplicate (already seen)?  & 
        \setlist{nolistsep}
        \begin{itemize}
            \item ``Yes'' or [BLANK]
        \end{itemize}\\
        \midrule
        Does this tweet violate Twitter's COVID-19 unapproved treatment policy?& 
        \setlist{nolistsep}
        \begin{itemize}
            \item Answer ``NA'' if previous answer was [BLANK], otherwise answer ``1'' - ``5'' based on the attached table (Table~\ref{tab:likert_descriptions})
        \end{itemize} \\
    \bottomrule
    \end{tabularx}
    \caption{Human evaluation question and guidelines provided to annotators during the COVID-19 policy violation verification task}
    \label{tab:guidelines_policy_violation}
\end{table*}

\paragraph{Identifying COVID-19 Policy Violation Tweets.}
Given a tweet with \textsc{SUPPORTING} stance towards the claim that ``\textit{treatment is effective in treating COVID-19}'', annotators assign a score based on a five-point Likert scale, with 5 corresponding to a clear violation of the policy and 1 representing a clear non-violation. Table~\ref{tab:likert_descriptions} shows the Likert score descriptions and Table~\ref{tab:guidelines_policy_violation} shows the annotation questions and guidelines as they appeared to annotators during the human-in-the-loop evaluation for this task.

\section{Annotation Interface for Stance Classification}
\label{appendix-anno}
\begin{figure}[ht]
    \centering
    \includegraphics[width=0.49\textwidth]{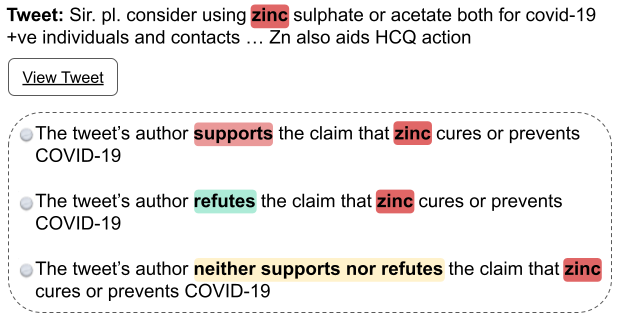}
    \caption{Amazon MTurk interface for stance annotation towards extracted claims in tweets.}
    \label{fig:annotation_task}
\end{figure}

Our stance data collection procedures (\S\ref{sec:stance}) on Amazon MTurk were approved by an ethics board. Before individuals were allowed to annotate data for our task, they were required to give consent by electronically signing off on the ethics statement found in Figure~\ref{fig:irb_statement} (some portions have been redacted for anonymity purposes). Note that all annotators were MTurk workers in the United States who had previously annotated $1000$ HITs with a pass-rate $\ge 95\%$.
Figure~\ref{fig:annotation_task} shows the interface used for collecting these stance annotations.

\begin{figure*}[t!]
    \centering
    \small
    \begin{tabularx}{0.92\textwidth}{p{0.92\linewidth}}
        \textit{``You are being asked to be a volunteer in a research study. The purpose of this study is to advance research on Computational Linguistics. The annotation form will take approximately 3 minutes to complete. You must be 18 years of age or older to participate. Your judgments will be used by researchers worldwide to help advance research on Computational Linguistics.  They will enable machine learning techniques to be applied to problems in natural language understanding. We will keep your personal information (Mechanical Turk ID, etc.) confidential. The risks involved are no greater than those involved in daily activities. We will comply with any applicable laws and regulations regarding confidentiality. To make sure that this research is being carried out in the proper way, the [REDACTED] may review study records. The [REDACTED] may also look at study records.  If you have any questions about the study, you may contact [REDACTED]. If you have any questions about your rights as a research subject, you may contact [REDACTED] at [REDACTED]. Thank you for participating in this study. By completing the online survey, you indicate your consent to be in the study. Subjects located in the EU are not allowed to join this study.''}

    \end{tabularx}
    \caption{Amazon MTurk ethics statement which was shown to annotators before stance labeling task}
    \label{fig:irb_statement}
\end{figure*}

\begin{table}[t!]
    \centering
    \footnotesize
    \begin{tabular}{cp{0.38\textwidth}}
        \toprule
        \textbf{Score} & \textbf{Description}\\
        \midrule
        1 & Clearly not in violation of Twitter's policy.\\
        2 & Probably not violating the policy, but does seem to suggest a questionable treatment may be effective. For example, the treatment is in clinical trials at the time the tweet was written, or the tweet does not make a strong claim about effectiveness.\\
        3 & Unclear whether or not this violates the policy. \\
        4 & Most likely violating Twitter's policy. Seems like the treatment is not effective based on official sources or reputable news organizations, and the tweet is making a relatively strong claim that the treatment is effective. \\
        5 & Clearly in violation of Twitter's policy. \\
        \bottomrule
    \end{tabular}
    \caption{Likert score descriptions that are presented to annotators.  These are used to evaluate whether tweets supporting a misinformation claim are in violation of Twitter's policies.}
    \label{tab:likert_descriptions}
\end{table}

\section{Approved Treatments}
\label{appendix-approved_treatments}
Table~\ref{tab:approved_treatements} shows the approved treatments that were used for filtering in \S\ref{subsec:ranking_trending}.
\begin{table}[t!]
    \centering
    \small
    \begin{tabularx}{0.48\textwidth}{@{} l X @{}}
        \toprule
         \textbf{Source} & \textbf{Treatments}\\
        \midrule
        CDC(*) & 
        Mask, 
        Face Mask,
        Social Distancing,
        Stay Home,
        Wash Hands,
        Hand Washing,
        Cover Coughs,
        Cover Sneezes\\
        \midrule
        New York Times(**) &
        Remdesivir, REGEN-COV, Bamlanivimab,
        Etesevimab,
        Sotrovimab,
        Dexamethasone,
        Prone positioning,
        Ventilators,
        Evusheld,
        Paxlovid,
        Molnupiravir,
        Lagevrio,
        Baricitinib,
        Olumiant,
        Tocilizumab,
        Actemra \\
    \bottomrule
    \end{tabularx}
    \caption{Approved COVID-19 treatments used in evaluation based on lists from the New York Times \protect\cite{nytimes_covid_treatements_2020} and the Centers for Disease Control \protect\cite{cdc_preventions_2021}.}
    \label{tab:approved_treatements}
\end{table}

\section{Implementation Details}
All experiments are performed with NVIDIA A40 GPUs. All hyperparameters are selected using a held-out development set.
\subsection{Claim Extraction Models}
\label{sec:appendix-claim}
Hyperparameters can be found in Table~\ref{tab:hyperparameters-claim}.
\paragraph{Sequence Tagging Models.}
We apply sequence-labeling models with a standard BIO tagging scheme (`B' and `I' tags are used to identify treatment tokens), similar to the named entity recognition method used by \citet{devlin-2019-bert}. 
We experiment with RoBERTa$_{large}$(354M)~\citep{liu2019roberta} and a domain-specific COVID-Twitter-BERT$_{large}$(345M) ~\citep{muller2020CTBERT}. 

\paragraph{Question Answering Models.}
We experiment with QA-based slot filling models~\citep{du-etal-2021-qa}, which model claim extraction as a SQuADv2.0 question-answering task~\citep{rajpurkar-etal-2018-squad2.0}.
We experiment with two approaches:
a span-prediction model that predicts start and end positions for answer spans in context using RoBERTa or CT-BERT as the encoder, in addition to a text-to-text model that generates answers using T5$_{large}$(770M)~\citep{raffel2020exploring,wang-lillis-2020-ucd}. 

\paragraph{Pre-training QA Models.}
We pre-train QA models on the SQuADv2.0 dataset for 2 epochs with a learning rate of 2e-5 and batch size of 16, followed by fine-tuning for claim extraction.

\begin{table}[t!]
    \centering
    \footnotesize
    \begin{tabular}{lcc}
         \toprule
          & Tagging/QA (non-T5) & QA (T5) \\
        \midrule
        learning rate & 1e-5,2e-5,3e-5 & 1e-4,2e-4,3e-4\\
        batch size & 8,16 & 8,16\\
        epoch & 50 & 10\\
         \bottomrule
    \end{tabular}
    \caption{Hyperparameters of claim extraction models.}
    \label{tab:hyperparameters-claim}
\end{table}

\subsection{Stance Classification Models}
\label{sec:appendix-stance}
Hyperparameters can be found in Table~\ref{tab:hyperparameters-stance}.
\paragraph{NBOW.}
We use an NBOW model and a relative position encoding (RPE)~\citep{shaw-etal-2018-self} to indicate the position of the claim in the input sentence. Specifically, the 1D RPE encodes the relative distance of each token to the extracted treatment. 
We then concatenate NBOW with the RPE embedding and pass the concatenation through one layer of a feed-forward neural network.

\paragraph{RoBERTa/CT-BERT.}
We finetune a RoBERTa$_{large}$ \cite{devlin-2019-bert} and COVID-Twitter-BERT (CT-BERT) \cite{muller2020CTBERT} using a linear classification layer. 
We use the best-performing model from \citet{soares2019re}, [{\sc Entity Markers - Entity Start}], which uses special tokens ${\tt [C_{start}], [C_{end}]}$ to mark treatment span in a sentence. 
The modified sentence is then fed into BERT/CT-BERT and the representation of the starting marker ${\tt [C_{start}]}$ is sent into a linear classification layer.

\begin{table}[t!]
    \centering
    \footnotesize
    \begin{tabular}{lccc}
         \toprule
          & NBOW & RoBERTa/CT-BERT\\
        \midrule
        learning rate & 1e-4,5e-5,1e-3,5e-3 & 8e-6,1e-5,3e-5\\
        batch size & 4,16 & 8,16\\
        epoch & 50 & 12\\
         \bottomrule
    \end{tabular}
    \caption{Hyperparameters of stance classification models.}
    \label{tab:hyperparameters-stance}
\end{table}

\section{Ranking Extracted Claims}
\label{sec:appendix-ranking}
To generate a list of treatments mentioned on a specific day sorted by trendiness or significance we cannot simply use the daily frequency counts of treatments mentioned, as more popular treatments will consistently be mentioned at a higher volume. In our evaluation dataset used in $\S$\ref{sec:extrinsic_human_eval}, for example, chloroquine and its variants are mentioned in a tweet with {\sc Supporting} stance approximately $11.5$ times per day on average while $56\%$ of total treatments encountered were mentioned less than one time in the period studied. Therefore, we require a method that takes into account the historical frequency of treatments to calculate the strength of the association between the trendiness of treatment and the date.

To do this, we use a one-tailed Fisher's Exact Test \cite{Fisher2010OnTI}, which has been shown to be effective in rare event detection applications in the domain of statistical natural language processing \citep{moore-2004-rare-events,johnson2007improving,ritter2011data}. 

To apply this test, we first calculate the hypergeometric probability, the probability of a particular distribution of treatment frequencies assuming independence between the treatment and the date. We define $T$ and $D$ as the events when a tweet's extracted treatment is $t$ and when a tweet is published on date $d$ respectively. Also, we let $C(X)$ be the observed frequency of event $X$ and $C(X,Y)$ be the joint frequency of event $X$ and event $Y$. Given these definitions, we can calculate the hypergeometric probability, $p_{T,D}$, as follows:
\begin{gather*}
\textstyle{p_{T,D} = \frac{C(T)!C(\neg T)!C(D)!C(\neg D)!}{N!C(T,D)!C(\neg T, D)!C(T,\neg D)! C(\neg T, \neg D)!}}
\end{gather*}
where $N$ is the sample size.

Given this formula of the hypergeometric probability for a distribution with a treatment $t$ and date $d$, we calculate the $p$-value of the test by summing the hypergeometric probabilities of this distribution and all more extreme distributions. In our case, more extreme distributions are hypothetical distributions where the joint frequency of a tweet with a specific treatment published on a specific date is greater than $C(T,D)$.

A treatment is flagged by our system if its $p$-value is less than the threshold or $\alpha$-value as set by the content moderator and it has not previously broken this threshold.

\section{Policy Violations Per Hour Calculation}
\label{sec:appendixcalculation}
Here we detail the calculation of the $124.2$ policy violations per hour statistic reported in \S\ref{sec:extrinsic_human_eval}. First, we calculate the total amount of time required by each of the phases of the human annotation:
\begin{enumerate}
    \item \textbf{Stage 1:} Detecting Misleading COVID-19 Treatments (\S\ref{subsec:early_detection})
    \begin{itemize}
        \item \# of claims = $300 \text{ claims}$
        \item time of verifying a claim $= 89.7 s/\text{claim}$
        \item time spent on claim annotation $= 300 * 89.7 s = 7.5 \text{ hours}$
    \end{itemize}
    \item \textbf{Stage 2:} Identifying COVID-19 Policy Violations (\S\ref{subsec:identifying_pv})
    \begin{itemize}
        \item \# of tweets mentioning an unapproved claim in the full batch $= 10246 \text{ claims}$
        \item time to annotate an individual tweet $= 16.1 s/\text{tweet}$
        \item estimated time spent on tweet annotation $= 10246*16.1 s = 45.8 \text{ hours}$
    \end{itemize}
\end{enumerate}
Next, we detail the calculation steps using these calculated times:
\begin{enumerate}
    \item Total annotation time $= 7.5 + 45.8 = 53.3 \text{ hours}$
    \item Estimated \#  of tweets (considering the full batch - see scores $4$ and $5$ bars in Figure \ref{fig:policy_violations}) containing policy violation $= 3151 + 3467 = 6618 \text{ tweets}$
    \item \# policy violated identified per hour of annotation time $= 6618 / 53.3 \text{ hours } = 124.2 \text{ tweets} / \text{hours}$
    \item total $\#$ tweets judged per hour of annotation time $= 10246 / 53.3 \text{ hours } = 192.2 \text{ tweets} / hour
$
\end{enumerate}
\end{document}